\documentclass[accepted]{uai2023} % for initial submission
\usepackage[american]{babel}
\usepackage{natbib} % has a nice set of citation styles and commands
\usepackage{mathtools} % amsmath with fixes and additions
\usepackage{booktabs} % commands to create good-looking tables
\usepackage{tikz} % nice language for creating drawings and diagrams
\usepackage{pifont}% http://ctan.org/pkg/pifont
\newcommand{\cmark}{\ding{51}}%
\newcommand{\xmark}{\ding{55}}%
\usepackage{array}
\newcolumntype{P}[1]{>{\centering\arraybackslash}p{#1}}
\usepackage{makecell}
\usepackage{commath}
\usepackage{graphicx}
\makeatletter

\newcommand{\Rmnum}[1]{\expandafter\@slowromancap\romannumeral #1@}
\makeatother
\setcitestyle{square, citesep={;}}
\title{Emotionally Enhanced Talking Face Generation}

\author[1]{Sahil Goyal}
\author[2,3]{Shagun Uppal}
\author[2,3]{Sarthak Bhagat}
\author[4]{Yi Yu}
\author[5]{Yifang Yin}
\author[2]{Rajiv Ratn Shah}
\affil[1]{%
    IIT, Roorkee\\
    India
}
\affil[2]{%
    IIIT, Delhi\\
    India
}
  \affil[3]{%
    Carnegie Mellon University\\
    USA
  }
\affil[4]{%
    National Institute of Informatics\\
    Japan
  }
\affil[5]{%
    A*STAR\\
    Singapore
  }
  
\begin{document}
\maketitle

\begin{abstract}
  % Lip synchronization and talking face generation have gained a specific interest from the research community with the advent and need of digital communication in different fields. 
  Several works have developed end-to-end pipelines for generating lip-synced talking faces with various real-world applications, such as teaching and language translation in videos. However, these prior works fail to create realistic-looking videos since they focus little on people's expressions and emotions. Moreover, these methods' effectiveness largely depends on the faces in the training dataset, which means they may not perform well on unseen faces. To mitigate this, we build a talking face generation framework conditioned on a categorical emotion to generate videos with appropriate expressions, making them more realistic and convincing. With a broad range of six emotions, i.e., \emph{happiness}, \emph{sadness}, \emph{fear}, \emph{anger}, \emph{disgust}, and \emph{neutral}, we show that our model can adapt to arbitrary identities, emotions, and languages. Our proposed framework is equipped with a user-friendly web interface with a real-time experience for talking face generation with emotions. We also conduct a user study for subjective evaluation of our interface's usability, design, and functionality. Project page: \href{https://midas.iiitd.edu.in/emo/}{\tt{https://midas.iiitd.edu.in/emo/}}
\end{abstract}

\section{Introduction}\label{sec:intro}
As the online consumption of digital video content increases, demand to generate short-duration videos has increased multi-fold. Researchers are working on constructing deep learning-based methods to generate high-quality videos capturing minute details with limited data and computational resources \citep{DBLP:journals/corr/abs-2103-00484, UPPAL2022149}. Talking face generation aims to create photo-realistic videos using visual (image or video) input and an audio source. These videos have applications in various domains, including digital animation, short tutorial creation, advertisements, etc. Most of the work done in this field predominantly focuses on either the quality of the videos \citep{10.1145/3072959.3073640, Zhang_2021_CVPR, yin2022styleheat}, or 
accurate sync of the audio and visual content of the videos \citep{Chen_2019_CVPR_Workshops,
% 10.1145/3306346.3323028, 
jamaludin2019you,thies2020neural}, but lack demonstrating relevant expressions, making the videos less realistic.

Predicting emotion from speech alone is difficult, requiring visual cues to understand or interpret the context. Visual emotions are critical factors that make these talking-face videos more realistic. Thus, those videos can be further employed for more practical purposes. This feature is often ignored or not modeled in most prior work in this area.
Earlier attempts \citep{DBLP:journals/corr/abs-1906-06337, https://doi.org/10.48550/arxiv.2007.08547} to infer facial emotions from audio have not been successful at accurately reproducing realistic animation and have struggled to control the facial expressions being depicted.
To incorporate emotion-conditioned expressions in the generated video, our work focuses on building a deep learning model to generate talking faces per the desired emotion category. We explicitly feed the selected emotion category as one hot vector to model and solely focus on enhancing the visual content to capture this expression appropriately. %However, we eventually plan to detect the emotion category directly from the input audio, thereby reducing another level of supervision. 
Our base architecture is similar to Wav2Lip \citep{prajwal2020lip}, and we additionally introduce an emotion encoder and emotion discriminator in our model to incorporate the emotion features. Our contributions are summarized as follows.
\begin{itemize}
    \item We propose a novel deep learning model that can generate photo-realistic lip-synced talking face videos, incorporating different emotions and associated expressions.
    \item We introduce a multimodal framework to generate lip-synced videos agnostic to any arbitrary identity, language, and emotion.
    \item We also develop a responsive web-based interface for real-time talking face generation with emotions.
\end{itemize}

\section{Related Work}
We review the work done in talking face generation and how human emotion is utilized in generating realistic talking face videos separately as follows.

\subsection{Talking Face Generation}
Several recent works focused on generating talking face videos using deep neural networks. \citet{wu2018reenactgan} proposed ReenactGAN for talking face generation using the face reenactment technique, which helped transfer the facial landmarks and expressions from a source video of an arbitrary person to the target identity. The landmark boundary encoding was extracted from an arbitrary person's video and mapped to the target person's video via a decoder.
% In this proposed method, the landmark boundary encoding is extracted from the arbitrary person's video. Then this landmark boundary encoding is mapped to the target person's video using a decoder. The quality of the proposed model was better than the CycleGAN technique. 
Some other works, such as \citep{Huang_2020_CVPR, Zhang_2020_CVPR}, also used facial landmark-based face reenactment techniques for generating video frames. \citet{chen2019hierarchical} used facial landmarks and a cascade GAN approach to generate desired videos. In this approach, the audio embedding was transferred to facial landmarks, which were then used to generate videos using a regression-based discriminator. \citet{Zhang_2021_ICCV} proposed Facial-GAN, which considered explicit face attributes like lip movements and implicit face attributes such as head pose and eye blink to generate high-quality video frames. Video-based methods that modified only the lip region of the face \citep{prajwal2020lip, thies2020neural, wen2020photorealistic, song2022everybody, wang2022attention} can generate high-quality talking face videos. They copied the upper half of the face from the input video to generate the target video and hence could not modify the facial expressions and emotions in the upper half of the face.
% \citep{ijcai2019-129} uses a Recurrent Adversarial Network to achieve temporal dependency across different video frames. This method achieves better video quality and increases lip synchronization accuracy by using a separate lip-reading discriminator. \citep{prajwal2020lip, wang2022attention} also uses a lip-sync discriminator to boost the accuracy of lip synchronization.
These works did not use human emotion in their models, one of the most critical explicit attributes that the model should incorporate to generate more realistic talking face videos.

\subsection{\textit{Emotional} Talking Face Generation}
% Most of the prior work in talking face generation only focuses on lip-synchronization and does not incorporate the speaker's emotion. However, in a practical scenario, the speaker's emotion plays a vital role in generating realistic videos.
Earlier methods \citep{DBLP:journals/corr/abs-1906-06337, https://doi.org/10.48550/arxiv.2007.08547} tried to infer facial emotions implicitly from audio. However, they have not been successful at accurately reproducing realistic animation and have struggled to control the facial expressions being depicted. In contrast, We explicitly feed the desired emotion category as the model input.

\begin{table}[t]
    \centering
    \begin{small}
    \caption {Recent audio-driven talking face generation methods. Most models that allow emotion control are image-based models (i.e., which use an identity image as an input along with speech utterance). (*) These methods do not explicitly learn the emotions but derive them implicitly from the audio input.}
    \label{tab:rw}
    \begin{tabular}{ |P{2.8cm}|P{1.2cm}|P{1.25cm}|P{1.1cm}|}
    \hline
    \textbf{Talking Face Generation Methods} & \textbf{Input (Image/ Video)} &\textbf{Arbitrary face}& \textbf{Emotion Synthesis}\\
    \hline
    \citet{das2020speech} & Image & \cmark & \xmark\\
    \hline
    MakeItTalk \citep{zhou2020makelttalk} & Image & \cmark & \xmark\\
    \hline
    \citet{zhang2021flow} & Image & \cmark & \xmark\\
    \hline
    \citet{wang2021audio2head} & Image & \cmark & \xmark\\
    \hline
    \citet{zhou2021pose} & Image & \cmark & \xmark\\
    % \hline
    % \citet{vougioukas2018end} & Image & \xmark & \xmark\\
    \hline
    \citet{thies2020neural} & Video & \cmark & \xmark\\
    \hline
    \citet{song2022everybody} & Video & \cmark & \xmark\\
    \hline
    Wav2Lip \citep{prajwal2020lip} & Video & \cmark & \xmark\\
    \hline
    % AttnWav2Lip \citep{wang2022attention} & Video & \cmark & \xmark\\
    % \hline
    \citet{wen2020photorealistic} & Video & \cmark & \xmark\\
    \Xhline{2\arrayrulewidth}
    \hline
    \citet{https://doi.org/10.48550/arxiv.2007.08547}* & Image & \xmark & \cmark\\
    \hline
    \citet{DBLP:journals/corr/abs-1906-06337}* & Image & \xmark & \cmark\\
    \hline
    \citet{9496264} & Image & \xmark & \cmark\\
    \hline
    MEAD \citep{wang2020mead} & Image & \xmark & \cmark\\
    \hline
    EVP \citep{Ji_2021_CVPR} & Video & \xmark & \cmark\\
    \hline
    \citet{sinha2022emotion} & Image & \cmark & \cmark\\
    \hline \hline
    \textbf{Ours} & \textbf{Video} & \cmark & \cmark\\
    \hline
    \end{tabular}
    \end{small}
\end{table}

Disentanglement \citep{Bengio2012RepresentationLA, higgins2017betavae, Mathieu2018DisentanglingDI, Shukla2019PrOSePO, Bhagat2020DisContSV, Bhagat2020DisentanglingMF}, which is defined as the process of extracting the underlying factors of variation in data into independent latent representations is a popular method to augment emotions in the generated videos.
\citet{Ji_2021_CVPR} proposed an \emph{Emotion Video Portraits} (EVP) algorithm to incorporate the emotion of the audio signal within the target video. Using a Cross-Reconstructed Emotion Disentanglement technique, they decomposed the audio input into a duration-dependent content feature and a duration-independent audio feature. With these two features, emotional facial landmarks were extracted. They introduced the Target-Adaptive Face Synthesis technique that adapted the inferred facial landmarks to the target video. However, they relied on intermediate global landmarks (or edge maps) to generate textures with emotions and on an additional Dynamic Time Warping \citep{berndt1994using} algorithm to develop their training data to enable cross-reconstructed training. Although they tried to learn emotion explicitly, the latent emotion representation was obtained by audio-emotion disentanglement. Hence, the disentanglement accuracy determined the control of the emotion, making it challenging to have flexible and fully independent control of the emotion. \citet{wang2020mead} proposed an emotional talking face generation method with explicit emotion control and MEAD dataset (a diverse emotional audio-visual dataset). Similar to our method, they used one-hot representation for emotion. However, they proposed a two-branch architecture, one branch for modifying only the upper half of the face based on emotions and the other for modifying only the lower half of the face using an LSTM \citep{hochreiter1997long}-based audio-to-landmarks module. This resulted in inconsistent and conflicting emotions on the face. So unlike the above-discussed methods, our work incorporates emotions into the whole face and uses an audio-independent emotion to generate the talking face videos. Also, EVP \citep{Ji_2021_CVPR} and MEAD \citep{wang2020mead} were involved in the training of target-specific texture models. Their work is based on single-identity generation. So unlike our model, they perform well only on the particular subject they are trained on and cannot adapt to arbitrary identities.

\citet{magnusson2021invertable} modified the architecture proposed in \citep{prajwal2020lip} to modify emotion using L1 reconstruction and pre-trained emotion objectives. However, their work suffered from several limitations. They did not modify the audio of the source video but retained the original one, which is not the case in most practical applications. 
% They ignored lip-synchronization and just focused on emotion transfer and identity preservation. 
In contrast, our model can choose arbitrary audio, ensuring lip synchronization accordingly.  Also, their model only modified emotion between specific pairs of emotions ({${\tt happiness}$}, {${\tt sadness}$}, and {${\tt neutral}$}), whereas our model has a broad range of six categorical emotions. Moreover, they trained separate models for each type of emotion transfer. In contrast, our single model can handle all kinds of emotion transfers.
% \citep{9496264} use an identity image, audio signal, and an explicit emotion condition to generate the video frames. Two separate discriminators, i.e., the frame and emotion discriminator, are used while generating the video frames.
% Expression-Tailored Generative Adversarial Network (ET-GAN) \citep{10.1145/3394171.3413844} used an expression video of an arbitrary user as a reference in addition to the identity image and audio source to generate the video. The facial landmarks are extracted from the expression video, then used to generate the video frames.
% GANimation \citep{10.1007/978-3-030-01249-6_50} provides an unsupervised technique to generate any number of discrete emotions using a GAN, conditioned on Action Units (AU) annotations, giving information about the facial movements that eventually define the human expression. Attention-based Identity Preserving Generative Adversarial Network (AIP-GAN) \citep{ali2020efficient} extracts the expression information from the input source
% image by calculating its facial landmarks using a supervised spatial and channel-wise
% attention module. 

Most models that allow emotion control are image-based models \citep{DBLP:journals/corr/abs-1906-06337, wang2020mead,https://doi.org/10.48550/arxiv.2007.08547, 9496264, sinha2022emotion} (i.e., which use an identity image as an input along with speech utterance), hence rendering only minor head movements and produce low-quality results. They cannot be used in real-world scenarios. Existing work in emotional talking face generation is limited (especially in the case of video-based models). To the best of our knowledge, this is one of the first studies in which the expression and emotion of a person are considered to generate lip synchronization and talking face generation from video input (See Table \ref{tab:rw}).

\section{Proposed Approach}
\begin{figure*}[t]
% \centerline{\includesvg[inkscapelatex=false, width=\textwidth]{final.svg}}
\includegraphics[width=\textwidth]{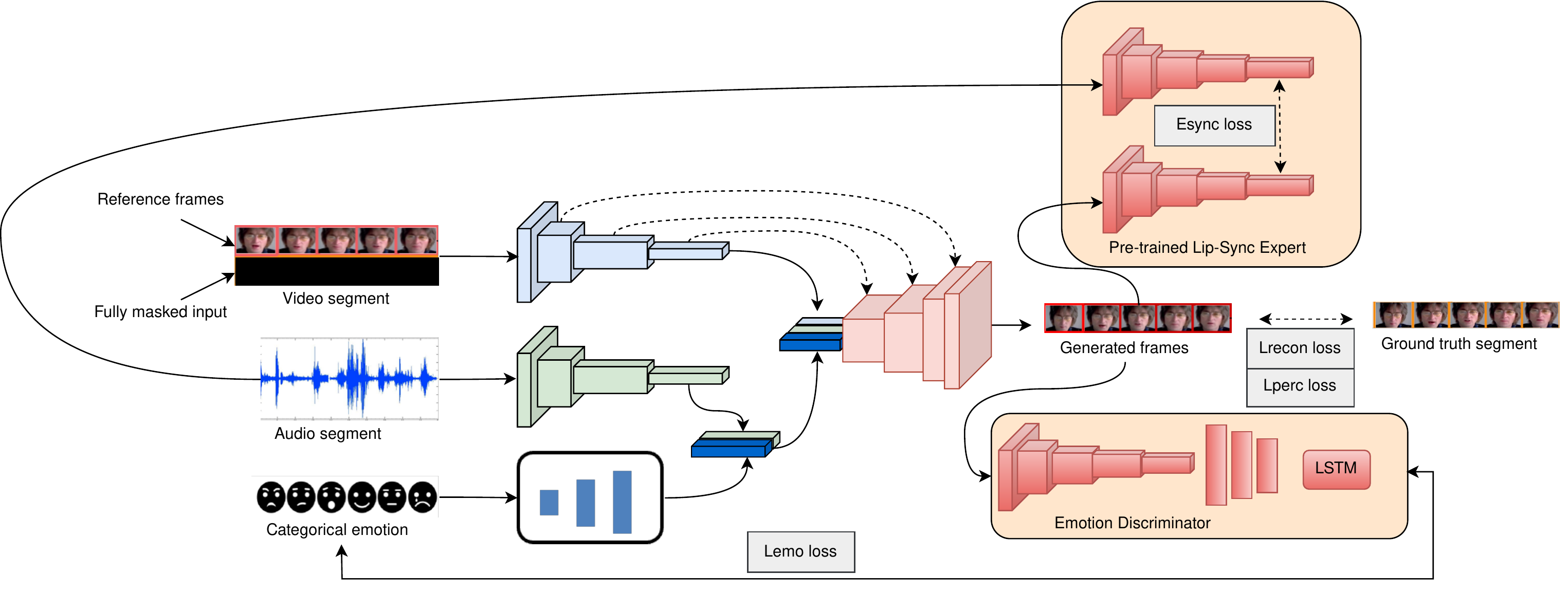}
\caption{We illustrate a video generation end-to-end network built upon base skeleton architecture. It accepts a continuous set of frames (fully masked) concatenated with reference frames, the Mel spectrogram form of a speech utterance, and a categorical emotion. Then concatenates their embeddings in a specific way as shown in this Figure to generate a lip-synced video rendered with the input emotion.}
\label{fig:main_model}
\end{figure*}

Our proposed framework aims at generating accurate lip-sync incorporated with appropriate emotions. In this section, we explain the different components of our method as illustrated in Figure \ref{fig:main_model}.

\subsection{Talking Face Generation}
\label{section:bm}
Our base skeleton is similar to \citep{prajwal2020lip}, which mainly emphasizes visual quality and accurate lip-sync generation. It comprises a generator that is a 2D-CNN encoder-decoder network to generate each frame independently. It broadly consists of three architectural blocks: (i) Face encoder, (ii) Audio encoder, and (iii) Decoder. Half-masked ground truth frames (the lower half is masked) concatenated with reference frames are used to generate the face embedding. 
% An example of half-masked input is shown in the supplementary material. 
Masked inputs ensure that the network gets the target-pose information but not the ground truth lip shape. The number of frames per input is set as $5$, i.e., $T=5$. Face embeddings from the face encoder and audio embeddings from the audio encoder are then passed through the decoder using skip connections (coming from outputs of layers of different resolutions of face encoder blocks) to generate the desired video. These skip connections ensure that fine-grained input facial features are preserved across deeper layers.
The generator is trained using a weighted combination of losses: (i) Reconstruction loss and (ii) Expert sync loss.

The generator is trained to minimize the difference between the generated frames $L_{g}$ and the ground truth frames $L_{G}$. For the above, $L1$ reconstruction loss is used.

\begin{equation}
    L_{recon} = \frac{1}{N}\sum_{i=1}^{N} \norm{L_{g}-L_{G}} _{1}
\end{equation}

A pre-trained \emph{expert lip-sync discriminator} is employed for accurate lip-syncing. A modified version of SyncNet \citep{chung2016out} is used for this task which is significantly deeper and contains residual skip connections \citep{he2016deep}.
When the whole model is trained, the generated frames (concatenated along the channel dimension) are input for the lip-sync discriminator 
\[ \{N,C,T,H,W\}  \equiv \{N,C*T,H,W\} \]
where $N$, $C$, $T$, $H$, and $W$ are batch size, number of channels, number of input frames, height, and width, respectively. Moreover, the lip-sync discriminator is not fine-tuned further on the generated frames. It is pre-trained as a classifier that determines whether an audio-video pair is synced.  Expert sync loss, which is essentially cosine similarity with binary cross-entropy loss, is used for optimizing the model weights:
\begin{equation}
    E_{sync} = \frac{1}{N}\sum_{i=1}^{N}-log(P_{sync}^{i}).
\end{equation}
To obtain $P_{sync}^{i}$, we compute a dot product between the ReLU-activated video and speech embeddings $v,s$ to get the probability of synchronization of an audio-video pair.
\begin{equation}
    P_{sync} = \frac{ v. s}{max(\norm{v}_{2}.\norm{s}_{2},\epsilon)}
\end{equation}

% \begin{figure}
%     \centering
%     \includegraphics[width=0.3\textwidth]{masked_input}
%     \caption{Masked Input, from \citep{prajwal2020lip}. The first row shows the reference frames, and the second row contains the half-masked frames. 
%     % Both sets of frames are continuous, i.e., have a temporal dependency.
%     }
%     \label{fig:masked_input}
% \end{figure}

\subsection{Emotion Capture in Talking Face Generation}
Current methods for generating talking face videos do not include sufficient information about the emotions and semantics of the subject. Visual emotions, along with visual quality and lip-syncing, are a significant part of any video to make it look natural. 
Also, inconsistent visual emotions can make it relatively easy for deepfake detectors to detect generated videos, as proposed in \citep{hosler2021deepfakes}.

Our approach, similar to \citep{9496264, Ji_2021_CVPR, sinha2022emotion}, ignores the emotion represented in the speech audio and conditions the video generation on an independent emotion label. Hence, this gives us more flexible control over the subject's emotions.

\subsubsection{Data Preprocessing}
In \citep{prajwal2020lip}, half-masked ground truth frames, along with reference frames, were used as the video input for the generator. So the task of the generator was to generate only masked lip-region to focus on accurate lip-sync generation. However, to incorporate the emotions, we use fully masked frames as input along with the reference frames because emotions are not only depicted by the lip region of a face; other regions of the face also depict them. As fully masked inputs do not provide additional information to the model, we expect similar results with only reference frames as input. 

\subsubsection{Data Augmentation (${\tt DA}$)}
\label{section:da}
We employed several data augmentation techniques on our input frames, such as random brightness contrast, random Gamma, channel shuffle, RGB shift, and Gaussian noise. The same augmentations were used in all the input frames to make the frames consistent in visual features like background color, contrast, luminance, brightness, etc. This helped us increase the training data and helped our model generalize over the different background settings.
% See Figure \ref{fig:data_aug} for an example.

% \begin{figure}[h]
%     \centering
%     \includegraphics[width=0.45\textwidth]{data_aug}
%     \caption{Augmented frame of an example of CREMA-D \citep{cao2014crema} dataset. The leftmost is the reference frame, followed by fully masked input, generated frame, and the ground truth frame.}
%     \label{fig:data_aug}
% \end{figure}

% \begin{figure*}[t]
%     \centering
%     \includegraphics[width=\textwidth]{final}
%     \caption{We illustrate a video generation end-to-end network build upon base skeleton architecture. It accepts a continuous set of frames (fully masked) concatenated with reference frames, the Mel spectrogram form of a speech utterance, and a categorical emotion concatenates their embeddings in a specific way as shown in the above Figure to generate a lip-synced video rendered with the input emotion.}
%     \label{fig:main_model}
% \end{figure*}

\subsubsection{Emotion Encoder}
We condition our video generation on categorical emotions. We assume $6$ basic emotion categories: {${\tt happiness}$},
{${\tt sadness}$}, {${\tt fear}$}, {${\tt anger}$}, {${\tt disgust}$}, and {${\tt neutral}$}. 
% However, this representation of emotions is not so precise as emotions are not as discrete as we have assumed. However, they are correlated or are a combination of two or more such discrete categories. Arousal-valence (AV) representation \citep{lewis2007neural,nicolaou2011continuous}, and hourglass of emotions \citep{cambria2012hourglass,susanto2020hourglass} are famous examples. Models using such a representation of emotions are generally more precise than models using categorical labels.\\
To encode these categorical emotions, we add an emotion encoder block to the generator described in Section ~\ref{section:bm}. We utilize a simple feed-forward neural network with Leaky ReLU activation as the emotion encoder. Emotion embedding obtained from the emotion encoder is similarly passed through the decoder as audio embedding is passed in Section ~\ref{section:bm}.

\subsubsection{Emotion Discriminator}
Our architecture for emotion discriminator is similar to that in \citep{9496264}. Each frame is passed through five 2D convolution layers. The convolution layers are as follows (representing the number of filters, kernel sizes, and strides, respectively): $(64, 3, 2)$, $(128, 3, 2)$, $(256, 3, 2)$, $(512, 3, 2)$, $(512, 3, 2)$, respectively. The output is then flattened and fed into a two-layer fully-connected network. The resulting sequence is fed into an LSTM \citep{hochreiter1997long} layer. The last time step of the LSTM layer's output is passed through a fully-connected layer that outputs probabilities for given categorical emotions.

Unlike the \emph{lip-sync discriminator} (refer Section \ref{section:bm}), input frames are concatenated across batch dimensions before passing through the emotion discriminator.
\[ \{N,C,T,H,W\}  \equiv \{N*T,C,H,W\} \]
% where $N, C, T, H, W$ are the batch size, number of channels, number of input frames, height, and width, respectively.
First, the emotion discriminator is pre-trained up to a few epochs; then, those weights are used as initialization to train it along with the generator. While updating the emotion discriminator in the final training, we compute the cross-entropy loss between the conditioned emotion label and the emotion label predicted by it for ground truth frames. In contrast, while updating the generator in the final training, we compute this loss between the conditioned emotion label and the emotion label predicted by the emotion discriminator for generated frames.

\subsection{Objective Functions}
We used multiple objective functions that emphasize different aspects of the generated video, such as visual quality, accurate lip-sync, and emotion rendering. 

\paragraph{Reconstruction Loss.}
The generator is trained to minimize the L1 reconstruction loss between the generated frames and the ground truth frames as described in Section ~\ref{section:bm}.

\paragraph{Penalizing Inaccurate Lip Generation.}
The generator is also trained to minimize the expert sync-loss $E_{sync}$ from the expert discriminator, which is cosine similarity with binary cross-entropy loss as described in Section ~\ref{section:bm}.
Remember that the pre-trained expert discriminator is not fine-tuned further during the training of the generator.

\paragraph{Perceptual Loss (${\tt PL}$).}
A pre-trained VGG-$19$ network \citep{simonyan2014very} is exploited to calculate the intermediate features of the layers from the ground truth videos and the generated videos. The mean-squared loss between these intermediate features is defined as the perceptual loss (${\tt PL}$) \citep{johnson2016perceptual}.

\paragraph{Emotion Discriminator Loss.}
The emotion discriminator is optimized using a cross-entropy loss calculated between the emotion class predicted by the emotion discriminator for generated frames and the conditioned emotion class. 
\begin{equation}
    L_{emo} = -\frac{1}{N}\sum_{i=1}^{N}y_{i}log(\hat{y}_{i})
\end{equation}
where, $N=6$ (here, $N$ signifies the number of emotion classes).

\paragraph{Combined Objective Function.}
The full objective function to train the generator:
\begin{equation}
\begin{split}
    L_{gen} = \alpha E_{sync} + \beta L_{perc} + \gamma L_{emo}\\ + (1-\alpha-\beta-\gamma) L_{recon}
\end{split}
\end{equation}
where, $\alpha, \beta, \gamma$ are the weights for the respective loss components.

\section{Experiments}
In this section, we discuss the dataset utilized, methods used for concatenating embeddings, and our experimental findings. Our code is available at \href{https://github.com/sahilg06/EmoGen}{\tt{https://github.com/sahilg06/EmoGen}}
% Noise encoder and implementation details are included in the supplementary material.

\subsection{Dataset}
To incorporate the emotions, a dataset with emotion labels is required, and according to our approach, it should fulfill the requirement of a single face in every frame of each clip.
Currently, only some such datasets are publicly available. 
% We explored various datasets such as CREMA-D \citep{cao2014crema}, RAVDESS \citep{livingstone2018ryerson}, etc.
We use CREMA-D for our purpose. Here are the main attributes of the dataset:
\begin{itemize}
    \item It contains $7442$ clips from $91$ actors ($48$ male and $43$ female).
    \item Actors spoke from a selection of $12$ sentences.
    \item Sentences were presented using one of the six emotions ({${\tt happiness}$}, {${\tt sadness}$}, {${\tt fear}$}, {${\tt anger}$}, {${\tt disgust}$}, {${\tt neutral}$}).
    \item The image resolution of the clips is $480 \times 360$.
\end{itemize}
We use $95\%$ as training data and $5\%$ as testing data. 
As mentioned in Section ~\ref{section:da}, data augmentation is also included to generalize our model better.

\subsection{Concatenating Methods}
\label{section:cm}
We try to concatenate the emotion embedding to video and speech embedding using two approaches:
\paragraph{End Concatenation (${\tt END}$).}
    We concatenate the emotion encoding at the final step with the video and audio encoding already concatenated. For this, we repeat the emotion $T=5$ (number of frames per input) along the first dimension. Then after passing through the emotion encoder, we get a latent representation of emotion which is then concatenated with already concatenated audio and face embeddings and is eventually passed through the final output block to get the generated frames of the video. 
        $$\underset{\text{face and audio embeddings}}{\underset{\text{Already concatenated }}{\{N*T,80,96,96\}}} + \underset{\text{Emotion embedding}}{\{N*T,1,96,96\}}$$
        $$ \equiv \underset{\text{Final embedding}}{\{N*T,81,96,96\}}$$
        $N, T$ are batch size and the number of input frames.
    Note that to concatenate the audio and video embeddings, we process them through face decoder blocks using skip connections (coming from outputs of layers of different resolutions of face/video encoder blocks).

 \paragraph{Sequential Concatenation (${\tt SEQ}$).}
    We concatenate the emotion encoding through skip connections similar to the audio encoding in Section \ref{section:bm}. We first concatenate the audio and emotion embedding. The concatenated embedding is processed through face decoder blocks of the generator using skip connections along with face embedding as shown in Figure \ref{fig:main_model}.

% \subsection{Noise Encoder}
% We introduce a noise encoder in the initial part of our model, along with a face, audio, and emotion encoder. A noise vector is drawn from the standard Gaussian distribution for each video frame. We process this sequence of noise vectors through a single layer of an LSTM \citep{hochreiter1997long} encoder to get noise embedding which is concatenated with the face embeddings. The motive for introducing this module is to account for some randomness, such as head movements and blinking of eyes, independent of the input data. We do not incorporate a noise encoder in the other experiments.

\begin{figure}[t]
    % \centerline{\includesvg[inkscapelatex=false,width=0.5\textwidth]{results.svg}}
    \includegraphics[width=0.5\textwidth]{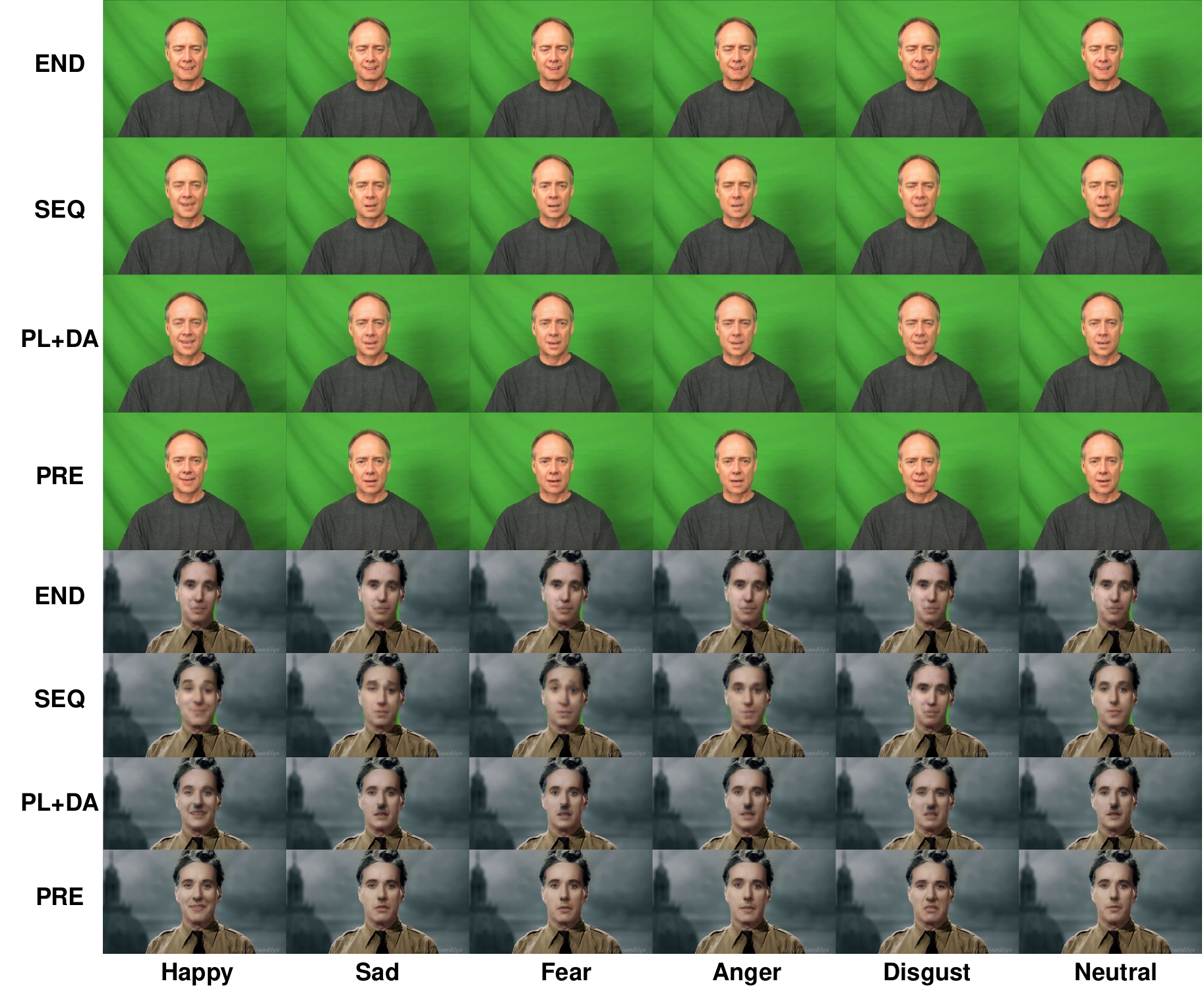}
    \caption{We generated videos for all six emotions and concatenated the specific frames from each. Each row represents an experiment mentioned in Section ~\ref{section:r}, and each column represents a particular emotion in all the experiments.}
    \label{fig:results}
\end{figure}

% \begin{figure*}
%     \centering
%     \includegraphics[width=\textwidth]{frame}
%     \caption{An example comparing generated frames of the baseline approach (Wav2Lip \citep{prajwal2020lip}) and our proposed approach using an \emph{arbitrary} identity with different emotion categories. Every fifth frame is shown in each row.}
%     \label{fig:frame}
% \end{figure*}

\subsection{Pre-training the Base Model (${\tt PRE}$)}
\label{section:pbm}
LRS2 \citep{afouras2018deep} is relatively larger than CREMA-D \citep{cao2014crema} and has more complex head poses, but it cannot be used for our modified model because it does not have categorical emotion labels. Hence, we try to pre-train the base model (that does not require emotion labels) on the LRS2 dataset and then use the face encoder block from the pre-trained model in two ways (as the architecture of the face encoder is the same in both the base model and the modified model):
\begin{itemize}
    \item Keeping the weights of the face encoder fixed while training the modified model.
    \item Using pre-trained weights of face encoder as initialization for training the modified model.
\end{itemize}
We also modify the base model to generate the whole face instead of only the lip region and then pre-train it.

\begin{figure*}[t]
% \centerline{\includesvg[inkscapelatex=false,width=\textwidth]{frame.svg}}
\includegraphics[width=1.3\textwidth]{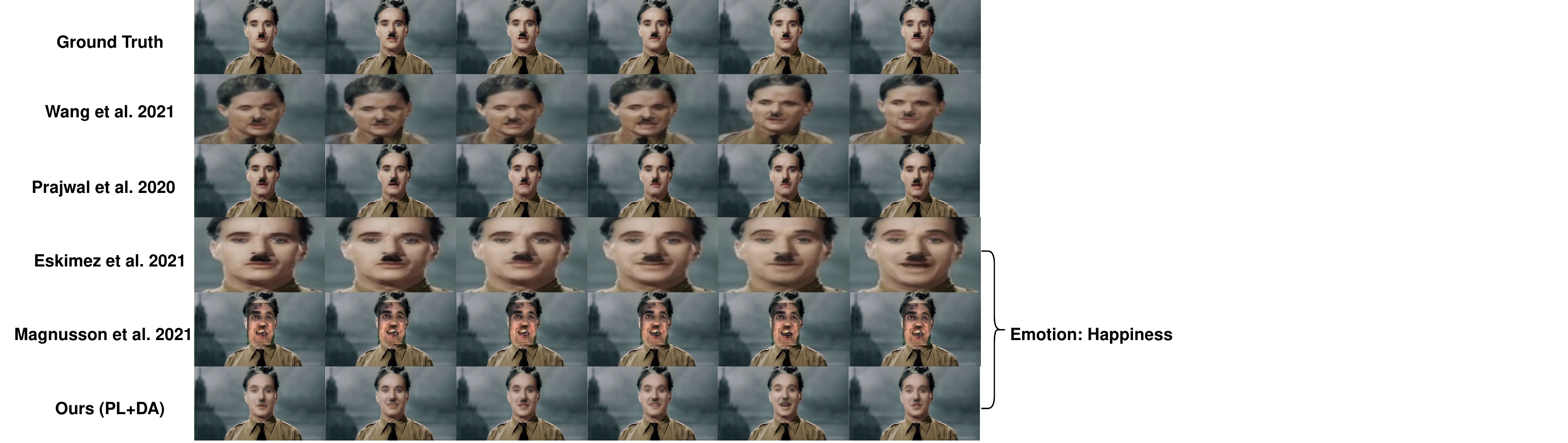}
\caption{An example comparing generated frames using an \emph{arbitrary} identity. Every fifth frame of the generated video is shown in each row. 
The first row corresponds to the ground truth video. Results corresponding to \citep{wang2021audio2head} (second row) and \citep{prajwal2020lip} (third row) do not involve emotion transfer.}
\label{fig:frame}
\end{figure*}

\section{Ablation Study}
\label{section:r}
We study the efficacy of our different experimental settings in this section.
\subsection{${\tt END}$ Concatenation}
See Section ~\ref{section:cm} for details of the ${\tt END}$ concatenation. We do not employ perceptual loss and data augmentation in this experiment. Although the sync quality is good, the visual quality and emotional rendering are unsatisfactory. See rows labeled ${\tt END}$ in Figure \ref{fig:results}. Moreover, some undesirable green background is present in the frames of the second example because all the training examples have a green screen in their background, so the model cannot generalize completely on other videos. Some arbitrary black dot artifacts are also visible on the generated frames. A possible explanation for the same could be that the one hot emotion vector is sparse. We repeat it for every frame and process this sparse vector formed through an emotion encoder to generate a large tensor, concatenating it to already concatenated audio and video embeddings to generate the required video. So the presence of large-sized sparse matrices in this approach results in black dot artifacts on the frames.

\subsection{${\tt SEQ}$ Concatenation}
See Section ~\ref{section:cm} for details of the ${\tt SEQ}$uential concatenation. This method improves the visual quality and emotional rendering to a large extent. Here, we do not employ perceptual loss or data augmentation. See rows labeled ${\tt SEQ}$ in Figure \ref{fig:results}. Emotion is rendered to some extent in the frames. The model still doesn't generalize, as a green background can be seen in the frames. However, those black dot artifacts disappear using the method ${\tt SEQ}$ because this approach reduces the size of the sparse matrices involved. This concatenation method is our preferred approach, and we conducted the following experiments using it.

% \paragraph{Evaluation of ${\tt SEQ}$.} 

\paragraph{Efficacy of including Perceptual Loss and Data Augmentation (${\tt PL}$+${\tt DA}$).} 
\label{exp3}
This approach is: ${\tt SEQ}$ $+$ Perceptual loss $+$ Data Augmentation.  See rows labelled ${\tt PL}$+${\tt DA}$ in Figure \ref{fig:results}. We observe the most satisfactory results under these experimental settings. Data augmentation solves the issue of a green background, aiding the model generalizing on videos other than training examples.
Also, penalizing the model with perceptual loss improves visual quality and emotion rendering.

\paragraph{Efficacy of pre-training the Base Model (${\tt PRE}$).}
\label{exp4}
This approach is basically: (${\tt PL}$+${\tt DA}$) + Pre-training. See Section ~\ref{section:pbm} for details of this experiment. 
% Note that perceptual loss, as well as data augmentation, are included in this approach.
See rows labeled ${\tt PRE}$ in Figure \ref{fig:results}.
The results show a slight improvement in the frames' visual quality, but a degradation in the temporal continuity of the generated frames is observed. Emotion rendering is comparable to ${\tt PL}$+${\tt DA}$.

\begin{table*}
    \centering
    \begin{small}
    \caption {Comparison of different approaches using Lip-Sync Error-Distance (\emph{LSE-D}), Lip-Sync Error-Confidence (\emph{LSE-C}), Emotion Classification Accuracy (\emph{EmoAcc}), and \emph{FID} score metrics.}
    \label{tab:quant_results}
    \begin{tabular}{|P{4.5cm}|P{1.5cm}|P{1.5cm}|P{1.5cm}|P{2.2cm}|P{1.5cm}|}
    \hline
    \textbf{Approach}&\textbf{Emotion} &\textbf{LSE-D} $\downarrow$ &\textbf{LSE-C} $\uparrow$ 
    &\textbf{EmoAcc $\uparrow$ (Top-1)}
    &\textbf{FID} $\downarrow$\\
    \hline
    Wav2Lip \citep{prajwal2020lip}& \xmark & \textbf{6.961} & 6.559 & - & 10.48\\ 
    \hline
    \citet{wang2021audio2head}& \xmark& 10.110 &4.976& - &72.81\\
    \hline
    \citet{DBLP:journals/corr/abs-1906-06337}&\cmark&- &- & 55.26 &71.12\\
    \hline
    \citet{9496264}&\cmark&10.244 &3.256 & 65.67& 79.11\\
    \hline
    \citet{sinha2022emotion}&\cmark&-&-& 75.02 &68.45\\
    \hline \hline
    {${\tt END}$} &\cmark& 7.754 & 6.369 & 21.48 & 15.68\\
    {${\tt SEQ}$} &\cmark& 7.464 &  6.201 & 71.51 & 14.32\\
    {${\tt PL+DA}$} &\cmark&  7.171 & \textbf{6.663} & \textbf{83.20} & 6.04\\ 
    {${\tt PRE}$}&\cmark& 7.946 & 6.053 &78.14 & \textbf{5.29}\\
    \hline
    \end{tabular}
    \end{small}
\end{table*}

\section{Qualitative Evaluation}
\label{section:qr}
We qualitatively compare our {${\tt PL+DA}$} approach with related works in this section.
We used an arbitrary identity sampled from the internet for qualitative comparison against other approaches. See Figure \ref{fig:frame} for results.
\cite{wang2021audio2head} and \cite{prajwal2020lip} did not involve emotion incorporation in their methodologies. Clearly, 
\begin{figure}[htp]
    \centering
    \includegraphics[width=0.272\textwidth]{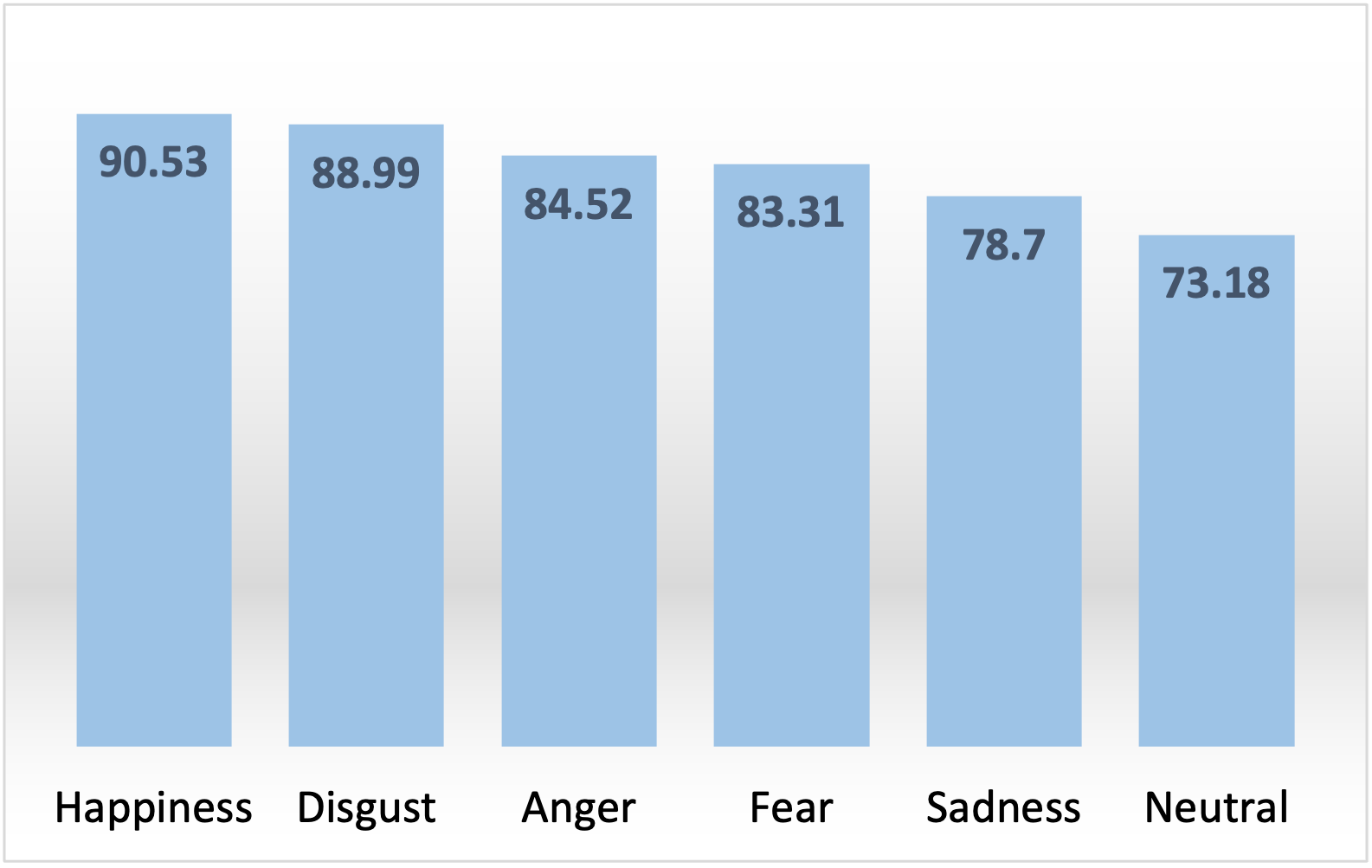}
    \caption{Emotion-wise accuracy for ${\tt PL}$+${\tt DA}$ approach.}
    \label{fig:emowise}
\end{figure} 
\citet{wang2021audio2head} failed to preserve the identity. For the emotional talking face generation methods \citep{9496264,magnusson2021invertable}, including our approach, we used {${\tt happiness}$} as the target emotion. \citet{9496264} could not effectively incorporate emotion into the generated frames. 
% As mentioned in the \emph{Related Work} section in the main paper, \citet{magnusson2021invertable} did not provide the freedom of choosing arbitrary audio. They did not modify the audio of the source video and ignored lip-syncing. They just focused on just emotion transfer of the given video. However, it is highly related to our work because of the similar base architecture and the similar task of emotion transfer.
\citet{magnusson2021invertable} 
% used a subset of the MEAD \citep{wang2020mead} dataset for training. In Figure \ref{fig:frame}, it 
clearly struggled to generalize on the arbitrary identity. An unwanted green background and many artifacts are present in the generated frames.

\section{Quantitative Results}
We evaluate our results against the state-of-the-art (SOTA) emotional talking face generation methods \citep{DBLP:journals/corr/abs-1906-06337,9496264, sinha2022emotion} for the essential attributes of a talking face, such as emotion incorporation, lip synchronization, and visual quality using the CREMA-D \citep{cao2014crema} dataset.
SOTA methods like Mead \citep{wang2020mead} and EVP \citep{Ji_2021_CVPR} are subject-specific. Their publicly available pre-trained models have been trained to perform well for a particular identity. We refrain from making quantitative comparisons with them to ensure fairness in evaluation. We also evaluate our results against the talking face generation methods, which do not incorporate emotions \citep{prajwal2020lip, wang2021audio2head} for lip-sync and visual quality. We summarize the quantitative results in Table \ref{tab:quant_results}. Note that the implementation code and pre-trained models for \citep{DBLP:journals/corr/abs-1906-06337, sinha2022emotion} are unavailable. Hence we only report the results mentioned in \citep{sinha2022emotion}. 

\subsection{Emotion Incorporation}
% We evaluated the emotion incorporation ability of our model quantitatively. 
We exploit an emotion classifier to evaluate the generated emotional talking face videos. We utilize the same architecture as the emotion discriminator in our main pipeline. We trained the classifier for the train split of the CREMA-D \citep{cao2014crema} dataset. We obtain an accuracy of more than $90\%$ on the test set of the CREMA-D dataset, indicating that our video-based emotion classification model can fairly evaluate the emotion incorporation ability of our model. The higher the emotion classification accuracy (\emph{EmoAcc}) of the video-based emotion classifier, the better the emotion incorporation ability of the model.

% For each video, we generate six new videos corresponding to the six emotions used in the approach and calculate average scores for those six videos in each experiment. 
As we are using arbitrary emotions to generate our videos, those arbitrary emotions can be exploited as ground truth labels for the classifier to evaluate our model. Table \ref{tab:quant_results} shows the best emotion classification accuracy (\emph{EmoAcc}) for all the approaches. Our ${\tt PL}$+${\tt DA}$ approach (defined in Section \ref{exp3}) gives the best \emph{EmoAcc} of $83.20\%$.
Emotion-wise accuracy for the ${\tt PL}$+${\tt DA}$ approach is depicted in Figure \ref{fig:emowise}. Note that \citep{prajwal2020lip, wang2021audio2head} do not incorporate emotions; therefore, we do not report their \emph{EmoAcc} in the Table \ref{tab:quant_results}.

\subsection{Sync Quality}
We use the metrics
% \footnote{Please refer to supplementary material for detailed calculation of the metrics \emph{LSE-C} and \emph{LSE-D}.}
\emph{LSE-C} and \emph{LSE-D}, proposed in \citep{prajwal2020lip} to evaluate the sync quality.
% \footnote{\href{https://github.com/joonson/syncnet_python}{https://github.com/joonson/syncnet\_python}}. 
% The first metric, Lip Sync Error - Distance (\emph{LSE-D}), calculated the distance between the audio and lip representations. The lower the \emph{LSE-D}, the higher the sync quality. The second metric, Lip Sync Error - Confidence (\emph{LSE-C}), calculates the average confidence score. Higher the \emph{LSE-C}, the higher the sync quality. For further details, please refer SyncNet \citep{chung2016out} paper.
% A sliding-window technique is utilized to calculate the metrics. For each video clip, multiple samples are extracted because there may be samples in which no one is speaking at that particular time. For each sample, the Euclidean distance between one 5-frame video feature and all the audio features in the $\pm$1 second range is calculated. Then those distances are averaged across all the samples. Out of all those average distances, the minimum one is defined as the Lip Sync Error - Distance (\emph{LSE-D}) because the correct offset is when the distance is minimum.
% The difference between the median and minimum (\emph{LSE-D}) of the average distances calculated above is defined as the Lip Sync Error - Confidence (\emph{LSE-C}). 
The lower the \emph{LSE-D}, the higher the sync quality. The higher the \emph{LSE-C}, the higher the sync quality.
We used the videos from the CREMA-D \citep{cao2014crema} dataset, but the audio inputs were randomly sampled from the internet in English and Hindi.
The scores of sync quality for all our experiments, including the related works, are shown in  Table \ref{tab:quant_results}. All our experiments ({${\tt END}$}, {${\tt SEQ}$}, {${\tt PL+DA}$}, {${\tt PRE}$}) have a sync quality comparable to our baseline (Wav2Lip \citep{prajwal2020lip}) and better than other related works, which means that adding emotion to the base model does not compromise the sync quality. Note that the metrics \emph{LSE-C} and \emph{LSE-D} are not reported for \citep{DBLP:journals/corr/abs-1906-06337, sinha2022emotion} in the Table \ref{tab:quant_results} because of the unavailability of their implementation code and pre-trained models.

% Code used...................
% import matplotlib.pyplot as plt
% import matplotlib.font_manager as fm
% categories = ['HAPPINESS', 'DISGUST', 'ANGER', 'FEAR', 'SADNESS', 'NEUTRAL']
% values = [90.53, 88.99, 84.52, 83.31, 78.70, 73.18]
% plt.bar(categories, values, color='#ADD8E6')
% plt.xlabel('Emotions', fontweight='bold', fontsize=13)
% plt.ylabel('EmoAcc',fontweight='bold', fontsize=13)
% font = fm.FontProperties(family='serif', size=11)
% plt.xticks(categories, categories, fontproperties=font)
% plt.subplots_adjust(right=1.1, top=1)
% for i, v in enumerate(values):
%     plt.text(i, v, str(v), ha='center', va='bottom', fontsize=11)
% plt.show()

% \begin{figure}[h]
%     \centering
%     \includegraphics[width=0.48\textwidth]{bar_graph_final}
%     \caption{Emotion-wise accuracy in decreasing order for ${\tt PL}$+${\tt DA}$ approach.}
%     \label{fig:emowise}
% \end{figure} 

\subsection{Visual Quality}
We use Fréchet Inception Distance (\emph{FID}) for evaluating the visual quality. Feature representations of the two sets of images are encoded using a pre-trained Inception network \citep{szegedy2015going}, and then Fréchet distance is calculated between the Gaussian distributions fitted to those representations. The \emph{FID} scores are shown in  Table \ref{tab:quant_results}. The \emph{FID} scores for all the approaches involving emotions are averaged over the six emotion categories. The \emph{FID} for our approach (involving emotion) is expected to be higher than the approaches not involving emotions \citep{prajwal2020lip, wang2021audio2head} because emotion incorporation, along with lip synchronization, requires more image manipulation than the ones involving only lip synchronization. The methods not incorporating emotions generate only the lower half region of the face, i.e., the lip region, whereas, for emotion incorporation, we generate the entire face. However, our visual quality improved significantly due to the addition of perceptual loss in ${\tt PL}$+${\tt DA}$ and ${\tt PRE}$ settings. The significant difference between ${\tt PL}$+${\tt DA}$ and ${\tt PRE}$ settings is additional knowledge gained by ${\tt PRE}$ through pre-training.
The ${\tt PRE}$ approach outperforms all other methods in \emph{FID}.

\begin{figure}[htp]
% \centerline{\includesvg[inkscapelatex=false,width=0.35\textwidth]{us.svg}}
\includegraphics[width=0.35\textwidth]{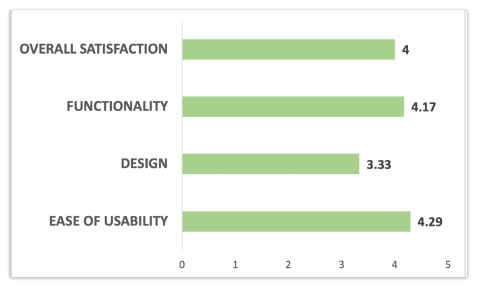}
\caption{User Study results for the web interface. The bar
plot depicts the average score on a scale of $0$ to $5$.}
\label{fig:us_eval}
\end{figure}

\section{User Study}
We conducted a user study through subjective evaluation to understand the user experience on our web interface. We surveyed a diverse group of $25$ users about their experience navigating and using the website. We asked them to rate the ease of usability, design, functionality, and overall experience on a scale of $0$ to $5$. 
%The higher the rating, the better the web interface.
%Additionally, we asked them for specific suggestions. We also provided them with a small illustration video explaining some basic steps to use the website. See Figure \ref{fig:user-study} for more details.
Figure \ref{fig:us_eval} shows the user study results. The user study results provided valuable insights into the strengths and weaknesses of our web interface.
% We identified areas where the design could be improved to make the user experience more intuitive and seamless. Moreover, these are some of the valuable specific suggestions given by the participants:
% \begin{itemize}
%     \item The website can be made more responsive.
%     \item Add a small description of how the target audio impacts the final output.
%     \item At the end, show the initial and final videos side by side. 
%     \item The text is hard to read because of the small font size, and the colors could be more appealing.
% \end{itemize}
% The feedback from the participants will allow us to improve the website a lot (especially the design of the website).
The feedback from the participants will enable us to improve the website, particularly its design significantly. 
% Technical details of the web interface are provided in the supplementary material.

\section{Conclusion}
In this work, we propose a novel end-to-end realistic video generation system that takes a set of continuous frames, a speech utterance, and a conditioned categorical emotion as input and generates an accurate lip-synced video incorporated with real emotions. We extend the problem of talking face generation by synthesizing expressions along with accurate lip movements. This work will surely lead to new directions in this field.

% \begin{contributions}
%     Briefly list author contributions. This is a nice way of making clear who did what and giving proper credit.
%     This section is optional.

%     H.~Q.~Bovik conceived the idea and wrote the paper.
%     Coauthor One created the code.
%     Coauthor Two created the figures.
% \end{contributions}

% \begin{acknowledgements} 
% \end{acknowledgements}

% References
\bibliography{main}

\begin{thebibliography}{49}
\providecommand{\natexlab}[1]{#1}
\providecommand{\url}[1]{\texttt{#1}}
\expandafter\ifx\csname urlstyle\endcsname\relax
  \providecommand{\doi}[1]{doi: #1}\else
  \providecommand{\doi}{doi: \begingroup \urlstyle{rm}\Url}\fi

\bibitem[Afouras et~al.(2018)Afouras, Chung, Senior, Vinyals, and
  Zisserman]{afouras2018deep}
Triantafyllos Afouras, Joon~Son Chung, Andrew Senior, Oriol Vinyals, and Andrew
  Zisserman.
\newblock Deep audio-visual speech recognition.
\newblock \emph{IEEE transactions on pattern analysis and machine
  intelligence}, 2018.

\bibitem[Bengio et~al.(2012)Bengio, Courville, and
  Vincent]{Bengio2012RepresentationLA}
Yoshua Bengio, Aaron~C. Courville, and Pascal Vincent.
\newblock Representation learning: A review and new perspectives.
\newblock \emph{IEEE Transactions on Pattern Analysis and Machine
  Intelligence}, 35:\penalty0 1798--1828, 2012.

\bibitem[Berndt and Clifford(1994)]{berndt1994using}
Donald~J Berndt and James Clifford.
\newblock Using dynamic time warping to find patterns in time series.
\newblock In \emph{KDD workshop}, volume~10, pages 359--370. Seattle, WA, USA:,
  1994.

\bibitem[Bhagat et~al.(2020{\natexlab{a}})Bhagat, Udandarao, and
  Uppal]{Bhagat2020DisContSV}
Sarthak Bhagat, Vishaal Udandarao, and Shagun Uppal.
\newblock Discont: Self-supervised visual attribute disentanglement using
  context vectors.
\newblock \emph{ArXiv}, abs/2006.05895, 2020{\natexlab{a}}.

\bibitem[Bhagat et~al.(2020{\natexlab{b}})Bhagat, Uppal, Yin, and
  Lim]{Bhagat2020DisentanglingMF}
Sarthak Bhagat, Shagun Uppal, Vivian Yin, and Nengli Lim.
\newblock Disentangling multiple features in video sequences using gaussian
  processes in variational autoencoders.
\newblock In \emph{European Conference on Computer Vision}, 2020{\natexlab{b}}.

\bibitem[Cao et~al.(2014)Cao, Cooper, Keutmann, Gur, Nenkova, and
  Verma]{cao2014crema}
Houwei Cao, David~G Cooper, Michael~K Keutmann, Ruben~C Gur, Ani Nenkova, and
  Ragini Verma.
\newblock Crema-d: Crowd-sourced emotional multimodal actors dataset.
\newblock \emph{IEEE transactions on affective computing}, 5\penalty0
  (4):\penalty0 377--390, 2014.

\bibitem[Chen et~al.(2019{\natexlab{a}})Chen, Maddox, Duan, and
  Xu]{chen2019hierarchical}
Lele Chen, Ross~K Maddox, Zhiyao Duan, and Chenliang Xu.
\newblock Hierarchical cross-modal talking face generation with dynamic
  pixel-wise loss.
\newblock In \emph{Proceedings of the IEEE/CVF Conference on Computer Vision
  and Pattern Recognition}, pages 7832--7841, 2019{\natexlab{a}}.

\bibitem[Chen et~al.(2019{\natexlab{b}})Chen, Zheng, Maddox, Duan, and
  Xu]{Chen_2019_CVPR_Workshops}
Lele Chen, Haitian Zheng, Ross Maddox, Zhiyao Duan, and Chenliang Xu.
\newblock Sound to visual: Hierarchical cross-modal talking face generation.
\newblock In \emph{Proceedings of the IEEE/CVF Conference on Computer Vision
  and Pattern Recognition (CVPR) Workshops}, June 2019{\natexlab{b}}.

\bibitem[Chen et~al.(2020)Chen, Cui, Liu, Li, Kou, Xu, and
  Xu]{https://doi.org/10.48550/arxiv.2007.08547}
Lele Chen, Guofeng Cui, Celong Liu, Zhong Li, Ziyi Kou, Yi~Xu, and Chenliang
  Xu.
\newblock Talking-head generation with rhythmic head motion, 2020.
\newblock URL \url{https://arxiv.org/abs/2007.08547}.

\bibitem[Chopra et~al.(2005)Chopra, Hadsell, and LeCun]{1467314}
S.~Chopra, R.~Hadsell, and Y.~LeCun.
\newblock Learning a similarity metric discriminatively, with application to
  face verification.
\newblock In \emph{2005 IEEE Computer Society Conference on Computer Vision and
  Pattern Recognition (CVPR'05)}, volume~1, pages 539--546 vol. 1, 2005.
\newblock \doi{10.1109/CVPR.2005.202}.

\bibitem[Chung and Zisserman(2016)]{chung2016out}
Joon~Son Chung and Andrew Zisserman.
\newblock Out of time: automated lip sync in the wild.
\newblock In \emph{Asian conference on computer vision}, pages 251--263.
  Springer, 2016.

\bibitem[Das et~al.(2020)Das, Biswas, Sinha, and Bhowmick]{das2020speech}
Dipanjan Das, Sandika Biswas, Sanjana Sinha, and Brojeshwar Bhowmick.
\newblock Speech-driven facial animation using cascaded gans for learning of
  motion and texture.
\newblock In \emph{European conference on computer vision}, pages 408--424.
  Springer, 2020.

\bibitem[Duchi et~al.(2011)Duchi, Hazan, and Singer]{duchi2011adaptive}
John Duchi, Elad Hazan, and Yoram Singer.
\newblock Adaptive subgradient methods for online learning and stochastic
  optimization.
\newblock \emph{Journal of machine learning research}, 12\penalty0 (7), 2011.

\bibitem[Eskimez et~al.(2021)Eskimez, Zhang, and Duan]{9496264}
Sefik~Emre Eskimez, You Zhang, and Zhiyao Duan.
\newblock Speech driven talking face generation from a single image and an
  emotion condition.
\newblock \emph{IEEE Transactions on Multimedia}, pages 1--1, 2021.
\newblock \doi{10.1109/TMM.2021.3099900}.

\bibitem[Goyal et~al.(2022)Goyal, Uppal, Bhagat, Goel, Mali, Yu, Yin, and
  Shah]{goyal2022emotional}
Sahil Goyal, Shagun Uppal, Sarthak Bhagat, Dhroov Goel, Sakshat Mali, Yi~Yu,
  Yifang Yin, and Rajiv~Ratn Shah.
\newblock Emotional talking faces: Making videos more expressive and realistic.
\newblock In \emph{Proceedings of the 4th ACM International Conference on
  Multimedia in Asia}, pages 1--3, 2022.

\bibitem[He et~al.(2016)He, Zhang, Ren, and Sun]{he2016deep}
Kaiming He, Xiangyu Zhang, Shaoqing Ren, and Jian Sun.
\newblock Deep residual learning for image recognition.
\newblock In \emph{Proceedings of the IEEE conference on computer vision and
  pattern recognition}, pages 770--778, 2016.

\bibitem[Higgins et~al.(2017)Higgins, Matthey, Pal, Burgess, Glorot, Botvinick,
  Mohamed, and Lerchner]{higgins2017betavae}
Irina Higgins, Loic Matthey, Arka Pal, Christopher Burgess, Xavier Glorot,
  Matthew Botvinick, Shakir Mohamed, and Alexander Lerchner.
\newblock beta-{VAE}: Learning basic visual concepts with a constrained
  variational framework.
\newblock In \emph{International Conference on Learning Representations}, 2017.
\newblock URL \url{https://openreview.net/forum?id=Sy2fzU9gl}.

\bibitem[Hochreiter and Schmidhuber(1997)]{hochreiter1997long}
Sepp Hochreiter and J{\"u}rgen Schmidhuber.
\newblock Long short-term memory.
\newblock \emph{Neural computation}, 9\penalty0 (8):\penalty0 1735--1780, 1997.

\bibitem[Hosler et~al.(2021)Hosler, Salvi, Murray, Antonacci, Bestagini,
  Tubaro, and Stamm]{hosler2021deepfakes}
Brian Hosler, Davide Salvi, Anthony Murray, Fabio Antonacci, Paolo Bestagini,
  Stefano Tubaro, and Matthew~C Stamm.
\newblock Do deepfakes feel emotions? a semantic approach to detecting
  deepfakes via emotional inconsistencies.
\newblock In \emph{Proceedings of the IEEE/CVF Conference on Computer Vision
  and Pattern Recognition}, pages 1013--1022, 2021.

\bibitem[Huang et~al.(2020)Huang, Yang, and Wang]{Huang_2020_CVPR}
Po-Hsiang Huang, Fu-En Yang, and Yu-Chiang~Frank Wang.
\newblock Learning identity-invariant motion representations for cross-id face
  reenactment.
\newblock In \emph{Proceedings of the IEEE/CVF Conference on Computer Vision
  and Pattern Recognition (CVPR)}, June 2020.

\bibitem[Jamaludin et~al.(2019)Jamaludin, Chung, and
  Zisserman]{jamaludin2019you}
Amir Jamaludin, Joon~Son Chung, and Andrew Zisserman.
\newblock You said that?: Synthesising talking faces from audio.
\newblock \emph{International Journal of Computer Vision}, 127\penalty0
  (11):\penalty0 1767--1779, 2019.

\bibitem[Ji et~al.(2021)Ji, Zhou, Wang, Wu, Loy, Cao, and Xu]{Ji_2021_CVPR}
Xinya Ji, Hang Zhou, Kaisiyuan Wang, Wayne Wu, Chen~Change Loy, Xun Cao, and
  Feng Xu.
\newblock Audio-driven emotional video portraits.
\newblock In \emph{Proceedings of the IEEE/CVF Conference on Computer Vision
  and Pattern Recognition (CVPR)}, pages 14080--14089, June 2021.

\bibitem[Johnson et~al.(2016)Johnson, Alahi, and
  Fei-Fei]{johnson2016perceptual}
Justin Johnson, Alexandre Alahi, and Li~Fei-Fei.
\newblock Perceptual losses for real-time style transfer and super-resolution.
\newblock In \emph{European conference on computer vision}, pages 694--711.
  Springer, 2016.

\bibitem[Magnusson et~al.(2021)Magnusson, Sankaranarayanan, and
  Lippman]{magnusson2021invertable}
Ian Magnusson, Aruna Sankaranarayanan, and Andrew Lippman.
\newblock Invertable frowns: Video-to-video facial emotion translation.
\newblock In \emph{Proceedings of the 1st Workshop on Synthetic
  Multimedia-Audiovisual Deepfake Generation and Detection}, pages 25--33,
  2021.

\bibitem[Masood et~al.(2021)Masood, Nawaz, Malik, Javed, and
  Irtaza]{DBLP:journals/corr/abs-2103-00484}
Momina Masood, Marriam Nawaz, Khalid~Mahmood Malik, Ali Javed, and Aun Irtaza.
\newblock Deepfakes generation and detection: State-of-the-art, open
  challenges, countermeasures, and way forward.
\newblock \emph{CoRR}, abs/2103.00484, 2021.
\newblock URL \url{https://arxiv.org/abs/2103.00484}.

\bibitem[Mathieu et~al.(2018)Mathieu, Rainforth, Siddharth, and
  Teh]{Mathieu2018DisentanglingDI}
Emile Mathieu, Tom Rainforth, N.~Siddharth, and Yee~Whye Teh.
\newblock Disentangling disentanglement in variational autoencoders.
\newblock In \emph{International Conference on Machine Learning}, 2018.

\bibitem[Prajwal et~al.(2020)Prajwal, Mukhopadhyay, Namboodiri, and
  Jawahar]{prajwal2020lip}
KR~Prajwal, Rudrabha Mukhopadhyay, Vinay~P Namboodiri, and CV~Jawahar.
\newblock A lip sync expert is all you need for speech to lip generation in the
  wild.
\newblock In \emph{Proceedings of the 28th ACM International Conference on
  Multimedia}, pages 484--492, 2020.

\bibitem[Shukla et~al.(2019)Shukla, Bhagat, Uppal, Anand, and
  Turaga]{Shukla2019PrOSePO}
Ankita Shukla, Sarthak Bhagat, Shagun Uppal, Saket Anand, and Pavan~K. Turaga.
\newblock Prose: Product of orthogonal spheres parameterization for
  disentangled representation learning.
\newblock \emph{ArXiv}, abs/1907.09554, 2019.

\bibitem[Simonyan and Zisserman(2014)]{simonyan2014very}
Karen Simonyan and Andrew Zisserman.
\newblock Very deep convolutional networks for large-scale image recognition.
\newblock \emph{arXiv preprint arXiv:1409.1556}, 2014.

\bibitem[Sinha et~al.(2022)Sinha, Biswas, Yadav, and
  Bhowmick]{sinha2022emotion}
Sanjana Sinha, Sandika Biswas, Ravindra Yadav, and Brojeshwar Bhowmick.
\newblock Emotion-controllable generalized talking face generation.
\newblock \emph{arXiv preprint arXiv:2205.01155}, 2022.

\bibitem[Song et~al.(2022)Song, Wu, Qian, He, and Loy]{song2022everybody}
Linsen Song, Wayne Wu, Chen Qian, Ran He, and Chen~Change Loy.
\newblock Everybody’s talkin’: Let me talk as you want.
\newblock \emph{IEEE Transactions on Information Forensics and Security},
  17:\penalty0 585--598, 2022.

\bibitem[Suwajanakorn et~al.(2017)Suwajanakorn, Seitz, and
  Kemelmacher-Shlizerman]{10.1145/3072959.3073640}
Supasorn Suwajanakorn, Steven~M. Seitz, and Ira Kemelmacher-Shlizerman.
\newblock Synthesizing obama: Learning lip sync from audio.
\newblock \emph{ACM Trans. Graph.}, 36\penalty0 (4), jul 2017.
\newblock ISSN 0730-0301.
\newblock \doi{10.1145/3072959.3073640}.
\newblock URL \url{https://doi.org/10.1145/3072959.3073640}.

\bibitem[Szegedy et~al.(2015)Szegedy, Liu, Jia, Sermanet, Reed, Anguelov,
  Erhan, Vanhoucke, and Rabinovich]{szegedy2015going}
Christian Szegedy, Wei Liu, Yangqing Jia, Pierre Sermanet, Scott Reed, Dragomir
  Anguelov, Dumitru Erhan, Vincent Vanhoucke, and Andrew Rabinovich.
\newblock Going deeper with convolutions.
\newblock In \emph{Proceedings of the IEEE conference on computer vision and
  pattern recognition}, pages 1--9, 2015.

\bibitem[Thies et~al.(2020)Thies, Elgharib, Tewari, Theobalt, and
  Nie{\ss}ner]{thies2020neural}
Justus Thies, Mohamed Elgharib, Ayush Tewari, Christian Theobalt, and Matthias
  Nie{\ss}ner.
\newblock Neural voice puppetry: Audio-driven facial reenactment.
\newblock In \emph{European conference on computer vision}, pages 716--731.
  Springer, 2020.

\bibitem[Uppal et~al.(2022)Uppal, Bhagat, Hazarika, Majumder, Poria,
  Zimmermann, and Zadeh]{UPPAL2022149}
Shagun Uppal, Sarthak Bhagat, Devamanyu Hazarika, Navonil Majumder, Soujanya
  Poria, Roger Zimmermann, and Amir Zadeh.
\newblock Multimodal research in vision and language: A review of current and
  emerging trends.
\newblock \emph{Information Fusion}, 77:\penalty0 149--171, 2022.
\newblock ISSN 1566-2535.
\newblock \doi{https://doi.org/10.1016/j.inffus.2021.07.009}.
\newblock URL
  \url{https://www.sciencedirect.com/science/article/pii/S1566253521001512}.

\bibitem[Van~der Maaten and Hinton(2008)]{van2008visualizing}
Laurens Van~der Maaten and Geoffrey Hinton.
\newblock Visualizing data using t-sne.
\newblock \emph{Journal of machine learning research}, 9\penalty0 (11), 2008.

\bibitem[Vougioukas et~al.(2019)Vougioukas, Petridis, and
  Pantic]{DBLP:journals/corr/abs-1906-06337}
Konstantinos Vougioukas, Stavros Petridis, and Maja Pantic.
\newblock Realistic speech-driven facial animation with gans.
\newblock \emph{CoRR}, abs/1906.06337, 2019.
\newblock URL \url{http://arxiv.org/abs/1906.06337}.

\bibitem[Wang et~al.(2022)Wang, Zhang, Xie, Huang, and Zha]{wang2022attention}
Ganglai Wang, Peng Zhang, Lei Xie, Wei Huang, and Yufei Zha.
\newblock Attention-based lip audio-visual synthesis for talking face
  generation in the wild.
\newblock \emph{arXiv preprint arXiv:2203.03984}, 2022.

\bibitem[Wang et~al.(2020)Wang, Wu, Song, Yang, Wu, Qian, He, Qiao, and
  Loy]{wang2020mead}
Kaisiyuan Wang, Qianyi Wu, Linsen Song, Zhuoqian Yang, Wayne Wu, Chen Qian, Ran
  He, Yu~Qiao, and Chen~Change Loy.
\newblock Mead: A large-scale audio-visual dataset for emotional talking-face
  generation.
\newblock In \emph{European Conference on Computer Vision}, pages 700--717.
  Springer, 2020.

\bibitem[Wang et~al.(2021)Wang, Li, Ding, Fan, and Yu]{wang2021audio2head}
Suzhen Wang, Lincheng Li, Yu~Ding, Changjie Fan, and Xin Yu.
\newblock Audio2head: Audio-driven one-shot talking-head generation with
  natural head motion.
\newblock \emph{arXiv preprint arXiv:2107.09293}, 2021.

\bibitem[Wen et~al.(2020)Wen, Wang, Richardt, Chen, and
  Hu]{wen2020photorealistic}
Xin Wen, Miao Wang, Christian Richardt, Ze-Yin Chen, and Shi-Min Hu.
\newblock Photorealistic audio-driven video portraits.
\newblock \emph{IEEE Transactions on Visualization and Computer Graphics},
  26\penalty0 (12):\penalty0 3457--3466, 2020.

\bibitem[Wu et~al.(2018)Wu, Zhang, Li, Qian, and Loy]{wu2018reenactgan}
Wayne Wu, Yunxuan Zhang, Cheng Li, Chen Qian, and Chen~Change Loy.
\newblock Reenactgan: Learning to reenact faces via boundary transfer.
\newblock In \emph{Proceedings of the European conference on computer vision
  (ECCV)}, pages 603--619, 2018.

\bibitem[Yin et~al.(2022)Yin, Zhang, Cun, Cao, Fan, Wang, Bai, Wu, Wang, and
  Yang]{yin2022styleheat}
Fei Yin, Yong Zhang, Xiaodong Cun, Mingdeng Cao, Yanbo Fan, Xuan Wang, Qingyan
  Bai, Baoyuan Wu, Jue Wang, and Yujiu Yang.
\newblock Styleheat: One-shot high-resolution editable talking face generation
  via pretrained stylegan.
\newblock \emph{arXiv preprint arXiv:2203.04036}, 2022.

\bibitem[Zhang et~al.(2021{\natexlab{a}})Zhang, Zhao, Huang, Zeng, Ni,
  Budagavi, and Guo]{Zhang_2021_ICCV}
Chenxu Zhang, Yifan Zhao, Yifei Huang, Ming Zeng, Saifeng Ni, Madhukar
  Budagavi, and Xiaohu Guo.
\newblock Facial: Synthesizing dynamic talking face with implicit attribute
  learning.
\newblock In \emph{Proceedings of the IEEE/CVF International Conference on
  Computer Vision (ICCV)}, pages 3867--3876, October 2021{\natexlab{a}}.

\bibitem[Zhang et~al.(2020)Zhang, Zeng, Wang, Pan, Liu, Liu, Ding, and
  Fan]{Zhang_2020_CVPR}
Jiangning Zhang, Xianfang Zeng, Mengmeng Wang, Yusu Pan, Liang Liu, Yong Liu,
  Yu~Ding, and Changjie Fan.
\newblock Freenet: Multi-identity face reenactment.
\newblock In \emph{Proceedings of the IEEE/CVF Conference on Computer Vision
  and Pattern Recognition (CVPR)}, June 2020.

\bibitem[Zhang et~al.(2021{\natexlab{b}})Zhang, Li, Ding, and
  Fan]{Zhang_2021_CVPR}
Zhimeng Zhang, Lincheng Li, Yu~Ding, and Changjie Fan.
\newblock Flow-guided one-shot talking face generation with a high-resolution
  audio-visual dataset.
\newblock In \emph{Proceedings of the IEEE/CVF Conference on Computer Vision
  and Pattern Recognition (CVPR)}, pages 3661--3670, June 2021{\natexlab{b}}.

\bibitem[Zhang et~al.(2021{\natexlab{c}})Zhang, Li, Ding, and
  Fan]{zhang2021flow}
Zhimeng Zhang, Lincheng Li, Yu~Ding, and Changjie Fan.
\newblock Flow-guided one-shot talking face generation with a high-resolution
  audio-visual dataset.
\newblock In \emph{Proceedings of the IEEE/CVF Conference on Computer Vision
  and Pattern Recognition}, pages 3661--3670, 2021{\natexlab{c}}.

\bibitem[Zhou et~al.(2021)Zhou, Sun, Wu, Loy, Wang, and Liu]{zhou2021pose}
Hang Zhou, Yasheng Sun, Wayne Wu, Chen~Change Loy, Xiaogang Wang, and Ziwei
  Liu.
\newblock Pose-controllable talking face generation by implicitly modularized
  audio-visual representation.
\newblock In \emph{Proceedings of the IEEE/CVF conference on computer vision
  and pattern recognition}, pages 4176--4186, 2021.

\bibitem[Zhou et~al.(2020)Zhou, Han, Shechtman, Echevarria, Kalogerakis, and
  Li]{zhou2020makelttalk}
Yang Zhou, Xintong Han, Eli Shechtman, Jose Echevarria, Evangelos Kalogerakis,
  and Dingzeyu Li.
\newblock Makelttalk: speaker-aware talking-head animation.
\newblock \emph{ACM Transactions on Graphics (TOG)}, 39\penalty0 (6):\penalty0
  1--15, 2020.

\end{thebibliography}

\appendix

\clearpage
\section{Appendix}

\begin{figure}[htp]
    \centering
    \includegraphics[width=0.4\textwidth]{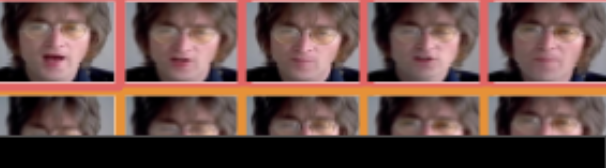}
    \caption{Masked Input, from Wav2Lip \cite{prajwal2020lip}. The first row shows the reference frames, and the second row contains the half-masked frames. Both sets of frames are continuous, i.e., they have a temporal dependency.}
    \label{fig:masked_input}
\end{figure}

\begin{figure}[htp]
    \centering
    \includegraphics[width=0.45\textwidth]{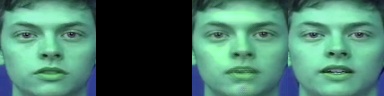}
    \caption{Augmented frame of an example of CREMA-D \cite{cao2014crema} dataset. The Leftmost is the reference frame, followed by fully masked input, generated frame, and the ground truth frame. All the frames are shown after applying data augmentation.}
    \label{fig:data_aug}
\end{figure} 

\subsection{Noise Encoder}
We introduce a noise encoder in the initial part of our model, along with a face, audio, and emotion encoder. A noise vector is drawn from the standard Gaussian distribution for each video frame. We process this sequence of noise vectors through a single layer of an LSTM \citep{hochreiter1997long} encoder to get noise embedding which is concatenated with the face embeddings. The motive for introducing this module is to account for randomnesses, such as head movements and eye blinking, independent of the input data. We do not incorporate a noise encoder in any of our experimental settings ({${\tt END}$}, {${\tt SEQ}$}, {${\tt PL+DA}$}, {${\tt PRE}$}).

\subsection{Implementation Details}
Adam optimizer \citep{duchi2011adaptive} is used for training all the networks with $\beta_{1}$ and $\beta_{2}$ as $0.5$ and $0.999$ respectively. The learning rate for updating the emotion discriminator and generator is $1e^{-6}$ and $1e^{-4}$, respectively. The full objective function of training the generator is
\begin{equation}
\begin{split}
    L_{gen} = \alpha E_{sync} + \beta L_{perc} + \gamma L_{emo}\\ + (1-\alpha-\beta-\gamma) L_{recon}
\end{split}
\end{equation}
where, $\alpha, \beta, \gamma$ are the weights for the respective loss components.
Constant $\alpha$ is set to $0$ initially and later updated to $0.03$ when the sync-loss on validation data becomes less than a predefined value. $\beta, \gamma$ are $0.01$ and $0.001$ respectively. Images are normalized between the $0$ and $1$ value range.  

By increasing the weight assigned to the emotion loss term, the model is able to more effectively incorporate emotions into its predictions at an earlier stage of the training process, but it comes at the cost of a slight reduction in reconstruction quality.

 \begin{figure}[htp]
    \centering
    \includegraphics[width=0.38\textwidth]{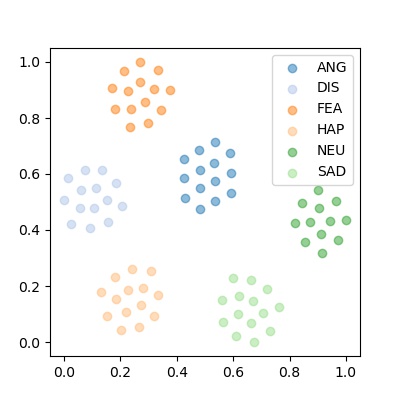}
    \caption{Visualization of the projected emotion embeddings. Each color represents a specific emotion.}
    \label{fig:tsne}
\end{figure}

\begin{figure}[htp]
% \centerline{\includesvg[inkscapelatex=false,width=0.45\textwidth]{demo.svg}}
\includegraphics[width=0.45\textwidth]{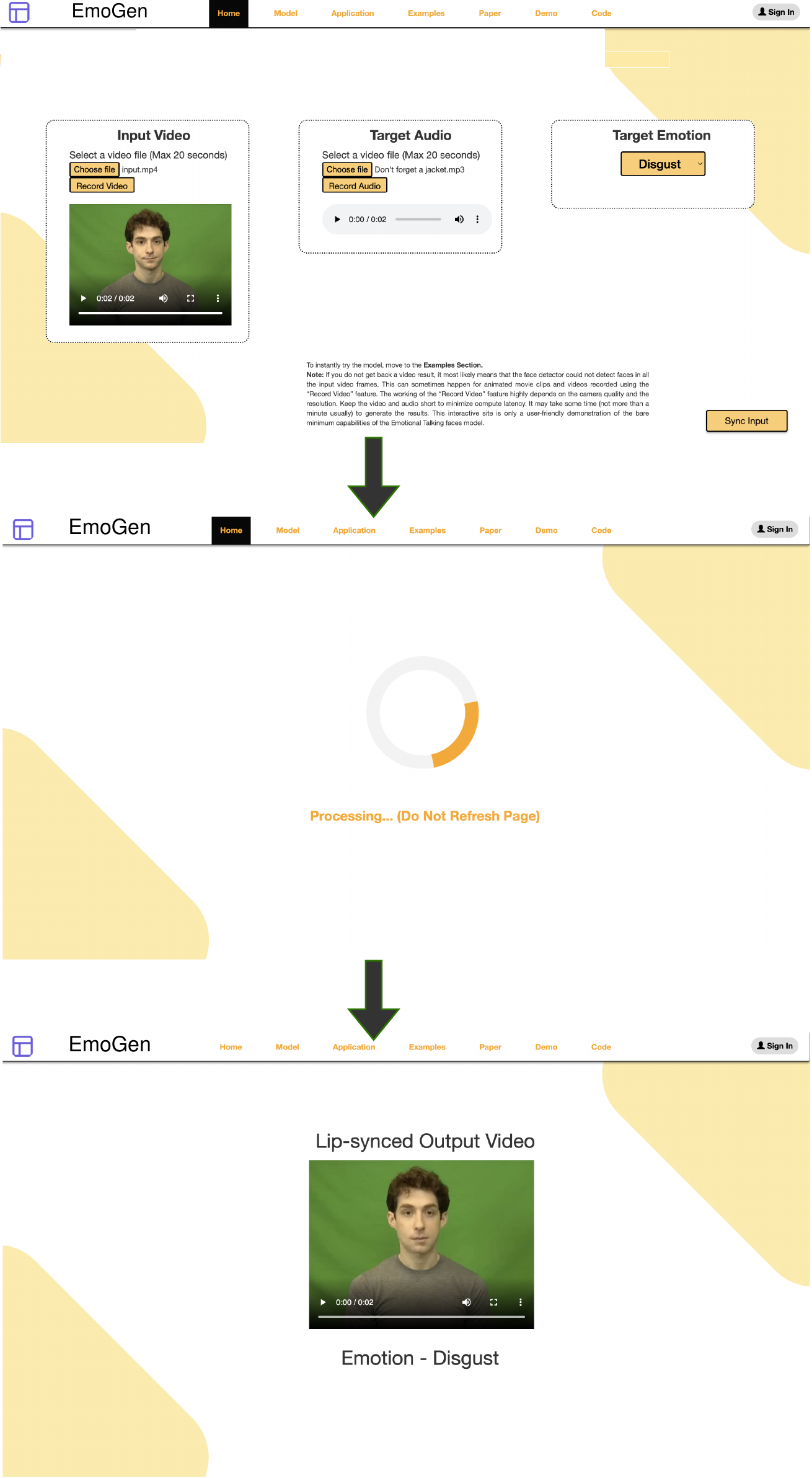}
\caption{Working of the demo website.}
\label{fig:demo}
\end{figure}

\begin{figure*}[htp]
% \centerline{\includesvg[inkscapelatex=false,width=\textwidth]{5frame_3.svg}}
\includegraphics[width=0.9\textwidth]{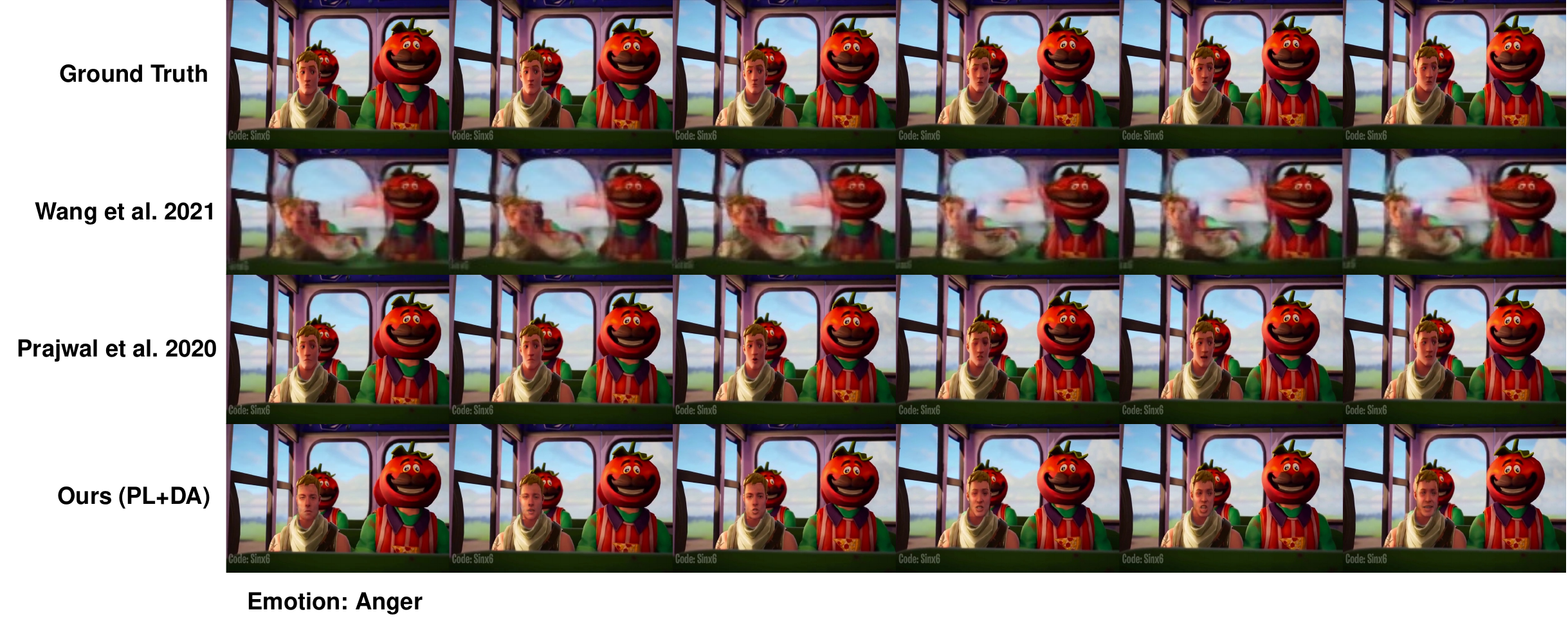}
\caption{An example comparing generated frames using a \textbf{cartoon subject} sampled from the internet. We chose this subject to evaluate the ability of different approaches to generalize to arbitrary identities. Every fifth frame of the generated video is shown in each row. \citet{wang2021audio2head} (second row) completely failed to generate any meaningful video and instead generated frames full of artifacts. \cite{9496264} was unsuccessful in detecting the relevant face from the video in the initial step and could thus not generate an emotional talking face video. Furthermore, \citet{magnusson2021invertable} cannot generate a video for \emph{anger} emotion.
In contrast, our approach {${\tt PL+DA}$} successfully detected the relevant face to generate the realistic frames and effectively conveyed the \emph{anger} emotion on the subject's face.}
\label{fig:cmp2}
\end{figure*}

\section{Visualizing the emotion embeddings}
We visualize the embeddings learned by our emotion encoder. We use \emph{t-SNE} \citep{van2008visualizing} algorithm to project the learned encodings to a $2$-dimensional space as shown in Figure \ref{fig:tsne}. We arbitrarily select ActorID $1011$ from the test split of the \citep{cao2014crema} dataset. We utilize all the videos of that actor for our purpose. We average the embeddings across the frames for each video. Each data point in Figure \ref{fig:tsne} represents averaged embeddings of a video of ActorID $1011$. Clusters formed for different emotions in Figure \ref{fig:tsne} show that our emotion encoder learns useful representations for the emotion.

\begin{figure*}[htp]
% \centerline{\includesvg[inkscapelatex=false,width=\textwidth]{5frame_2.svg}}
\includegraphics[width=1.\textwidth]{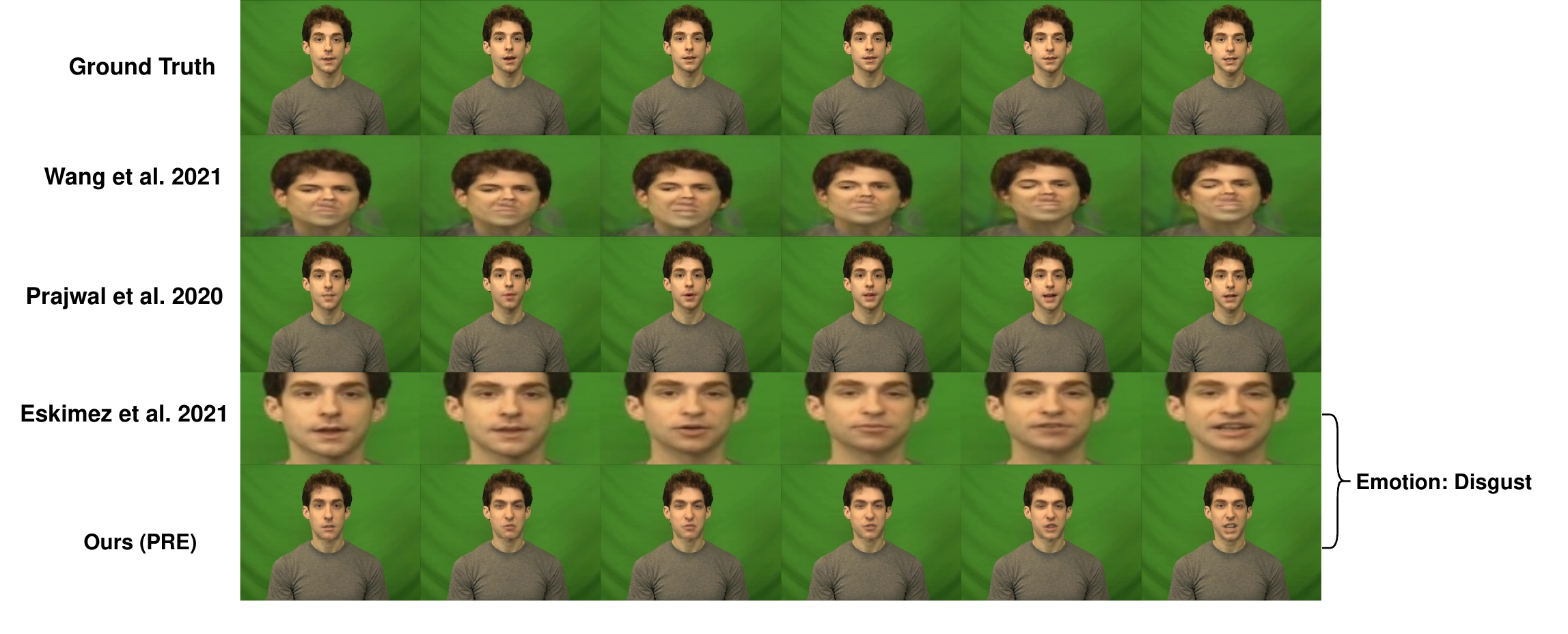}
\caption{An example comparing generated frames using a subject from the test dataset of CREMA-D \citep{cao2014crema}. Every fifth frame of the generated video is shown in each row. The top row corresponds to the ground truth video. Our baseline \citep{prajwal2020lip} (third row) generated realistic frames but cannot incorporate emotions. \citet{wang2021audio2head} (second row) again failed to preserve the subject's identity, resulting in non-human-like faces. \cite{9496264} (fourth row) could not effectively synthesize the \emph{disgust} emotion. \citet{magnusson2021invertable} involves only three emotions (\emph{happiness}, \emph{sadness}, \emph{neutral}). It cannot generate video for \emph{disgust} emotion.
In contrast, our approach {${\tt PRE}$} was able to generate realistic frames that accurately depicted the \emph{disgust} emotion on the subject's face.}
\label{fig:cmp2}
\end{figure*}

\section{Calculating \emph{LSE-C} and \emph{LSE-D}}
Pre-trained SyncNet released by \citep{chung2016out} is utilized to measure the lip-sync error between the generated frames and the randomly chosen speech segment. This SyncNet differs from the \emph{expert lip-sync discriminator} we have used in training. Its architecture is based on Siamese networks \citep{1467314} and is trained on a public dataset (derived from the BBC videos) using contrastive loss. The pre-trained model is available publicly\footnote{\href{https://github.com/joonson/syncnet_python}{https://github.com/joonson/syncnet\_python}}.

A sliding-window technique is utilized to calculate the \emph{LSE-C} and \emph{LSE-D} metrics. For each video clip, multiple samples are extracted because there may be samples in which no one is speaking at that particular time. The Euclidean distance between one 5-frame video feature and all the audio features in the $\pm$1 second range is calculated for each sample. Then those distances are averaged across all the samples. Out of all those average distances, the minimum one is defined as the Lip Sync Error - Distance (\emph{LSE-D}) because the correct offset is when the distance is minimum.
The difference between the median and minimum (\emph{LSE-D}) of the average distances calculated above is defined as the Lip Sync Error - Confidence (\emph{LSE-C}). 

\section{Web Interface}
Our proposed framework includes a user-friendly web interface \cite{goyal2022emotional} that allows users to generate talking faces with emotions using the model with ${\tt PL}$+${\tt DA}$ settings. Currently, the model uses an NVIDIA TITAN Xp GPU for inference.
FastAPI (Python Framework) is used for the backend development of the interface, which handles all the API requests. HTML, CSS, and Javascript are used for front-end development. All the clients' requests are sent to the backend via Javascript using a fetch call. Request details are sent in JSON format. The website
\footnote{\href{https://midas.iiitd.edu.in/emo/}{https://midas.iiitd.edu.in/emo/}} 
is hosted on HTTPS to address security issues. The website is super-easy to use, as illustrated in Figure \ref{fig:demo}. Following are some basic steps to use the demo website:
\begin{itemize}
    \item Before using the interface, read the instructions on the home page.
    \item Choose an arbitrary video, audio, and emotion as inputs. You can also use the recording feature for video and audio inputs. Then press the "Sync Input" button (located at the bottom right of the home page).
    \item After a $20$ to $30$ seconds wait, the emotionally enhanced and lip-synced talking face video will be ready.
\end{itemize}

\section{Limitations and Future Work}
Our approach, however, is limited by the availability of datasets with categorical emotion labels that are long enough and have exactly one face in each frame. Our current approach does not allow the use of datasets with multiple faces in a single frame, and the short datasets do not allow the model to generalize effectively. CREMA-D \citep{cao2014crema} contains relatively simple videos (with only a straight head pose). We can find or collect a better dataset for future work. It should be long enough to make the model generalize better and have videos with different head poses. One such potential dataset is MEAD \citep{wang2020mead}.

Various further improvements can be included in future work.
Some better masking methods can be explored to mask the ground truth frames (such as masking using a convex hull). Different ways to enforce the input emotion on the final audio can be examined, such as using an additional loss function. For evaluating the emotion rendering of the model, deepfake detectors that detect deepfakes based on inconsistency in emotions can be used. Also, some more relevant metric than \emph{FID} score is required to access the visual quality in the case of emotion incorporation because emotion rendering leads to more significant changes in the face compared to just lip synchronization. 

\section{Ethical Use}
Synthetic video generation has many potential applications, including entertainment, education, and marketing. However, their use also raises ethical concerns that must be carefully considered. Talking face generation videos may spread misinformation or propaganda or impersonate individuals for fraudulent or malicious purposes. It can lead to reputation damage and emotional distress. As these videos become more sophisticated and difficult to detect, it becomes increasingly challenging to distinguish real from fake content. This undermines the integrity of the media. Given these risks, it is essential to consider how synthetic media can be regulated or controlled to minimize their negative consequences. One possibility is developing high-quality algorithms or tools that detect and flag synthetic content. Another approach is establishing legal frameworks or guidelines that outline the acceptable uses of talking face generation videos and penalties for misuse. 

Finally, it is crucial to recognize that the creators and users of talking face generation videos are responsible for ensuring that they are used ethically, which includes considering the potential impacts of their work on others and taking steps to minimize any negative consequences. It also involves being transparent about synthetic media and clearly labeling content as manipulated when appropriate.
\end{document}